\documentclass[letterpaper, 10 pt, conference]{ieeeconf}

\usepackage{cite}
\usepackage{amsmath,amssymb,amsfonts}
\usepackage{algorithmic}
\usepackage{graphicx}
\usepackage{textcomp}
\usepackage{hyperref}
\usepackage{wrapfig}
\PassOptionsToPackage{usenames,dvipsnames}{xcolor}
\usepackage{forest, tikz}
\usepackage{mathptmx}

\usepackage{amsthm}
\usepackage{adigraph}
\usepackage{subcaption}
\usetikzlibrary{positioning, decorations.markings}
\usepackage{dirtytalk}
\usepackage{flushend}
\newtheorem{theorem}{Theorem}
\newtheorem{definition}{Definition}
\newtheorem{lemma}{Lemma}
\newtheorem{assumption}{Assumption}
\newtheorem{problem}{Problem}
\newtheorem{implementation}{Implementation}
\newcommand{\bt}{\mathcal{T}}

\IEEEoverridecommandlockouts

\begin{document}
\title{Adding Neural Network Controllers to Behavior Trees \\without Destroying Performance Guarantees}
\author{Christopher Iliffe Sprague \and Petter \"Ogren
\thanks{This  work  was  supported  by SSF  through  the  Swedish  Maritime Robotics Centre (SMaRC) (IRC15-0046).}
\thanks{C. I. Sprague and P. \"Ogren are with the Robotics, Perception and Learning Lab., School of Electrical Engineering and Computer Science, 
Royal Institute of Technology (KTH), SE-100 44 Stockholm, Sweden (e-mail: \href{mailto:sprague@kth.se}{sprague@kth.se}).}%
}

\maketitle
\thispagestyle{empty}

\begin{abstract}
    In this paper, we show how Behavior Trees that have performance guarantees, in terms of safety and goal convergence, can be extended with components that were designed using machine learning, without destroying those performance guarantees.

    Machine learning approaches such as reinforcement learning or learning from demonstration can be very appealing to AI designers that want efficient and realistic behaviors in their agents. However, those algorithms seldom provide guarantees for solving the given task in all different situations while keeping the agent safe. Instead, such guarantees are often easier to find for manually designed model-based approaches. In this paper we exploit the modularity of behavior trees to extend a given design with an efficient, but possibly unreliable, machine learning component in a way that preserves the guarantees.
    The approach is illustrated with an inverted pendulum example.
\end{abstract}


\section{Introduction}
\label{sec:intro}

Over the last decade, Behavior Trees (BTs) have become a very important tool in the design of decision structures, and are widely appreciated for their modularity and reactivity, see \cite{champandard2010behavior,colledanchise2017behavior,sprague2021continuous}, and the over 150 papers cited in \cite{iovino2020survey}. At the same time, machine learning (ML) approaches have continued to show remarkable success, \cite{Mnih2015,silver2016mastering,justesen2019deep}. However, one problem with ML approaches is that they seldom provide guarantees for ending up at the desired state, or for avoiding some unsafe states that might harm the agent. 
In this paper, we will show how, and when, the BT design of Figure~\ref{fig:cyclical_bt} can combine the safety guarantees of an safety controller with the performance guarantees of a model-based controller and the efficiency of a data-driven controller.

The basic idea is very straightforward and relies on the modularity, BTs where proven to be optimally modular in \cite{biggar2020modularity}, provided by the BT structure. Note that the rest of the BT can be arbitrarily complex, but we focus on what is going on when the given subtree of interest is executed. Looking at Fig.~\ref{fig:cyclical_bt}, the first priority is safety, and whenever the safety constraint is at risk of being violated we invoke the Safety controller, this might e.g., correspond to moving away from the edge of a cliff. If safety is ok, the BT checks if the current running cost, e.g. execution time, is ok, i.e. if there is reason to believe that the data-driven subtree is not going to complete the task in time, or at all. If the execution time is not ok, the previously designed model-based controller is invoked. Finally, if both of the conditions above are satisfied, we allow the data-driven subtree to be executed.

\begin{figure}[t]
    \footnotesize
    \begin{subfigure}[c]{0.35\linewidth}
        \centering
        \begin{forest}
            for tree={
                minimum height=2em,
                minimum width=2em,
                inner sep=0.8pt
            }
            [Rest of BT, edge=dashed, draw
            [, edge=dashed]
            [$\begin{array}{c}
                \text{Model-based} \\
                \text{controller}
            \end{array}$, draw]
            [, edge=dashed]
            ]
        \end{forest}
    \end{subfigure}
    \begin{subfigure}[c]{0.35\linewidth}
        \centering
        \begin{forest}
            for tree={
                minimum height=2em,
                minimum width=2em,
                inner sep=0.8pt
            }
            [Rest of BT, edge=dashed, draw
            [, edge=dashed]
            [$\rightarrow$, draw, edge=dashed,tikz={\node[left=2pt of .west] {$0$};}
                [$\begin{array}{c}
                    \text{Safety} \\
                    \text{controller}
                \end{array}$, draw, tikz={\node[above=2pt of .north] {$1$};}],
                [$?$, draw, tikz={\node[below=2pt of .south] {$2$};}
                    [$\begin{array}{c}
                        \text{Running cost} \\
                        \text{too high?}
                    \end{array}$, draw,ellipse, tikz={\node[below=2pt of .south] {$3$};}],
                    [$\begin{array}{c}
                        \text{Data-driven} \\
                        \text{controller}
                    \end{array}$, draw, tikz={\node[below=2pt of .south] {$4$};}]
                ]
                [$\begin{array}{c}
                    \text{Model-based} \\
                    \text{controller}
                \end{array}$, draw, tikz={\node[right=2pt of .east] {$5$};}],
            ],
            [, edge=dashed]
            ]
        \end{forest}
    \end{subfigure}
    \caption{A Model-based controller (left) is replaced by a subtree including both the model-based controller and the data-driven controller (right).}
    \label{fig:cyclical_bt}
\end{figure}

Note that switching between subtrees like this might induce undesired behaviors where one subtree counteracts another one, therefore, the rest of this paper is devoted to finding explicit formal conditions for when  the approach outlined above will indeed provide the desired guarantees, building upon the theoretical tools proposed in \cite{colledanchise2017behavior,sprague2021continuous},
and illustrating the approach with the commonly known example of an inverted pendulum swingup problem.

The main contribution of this paper is that we show how to add a data-driven controller to an \emph{existing} BT design, without destroying performance guarantees.

The outline of the paper is as follows.
In Section~\ref{sec:related_work} and \ref{sec:background} we present related work and background material. Then, in Section~\ref{sec:main_result} we formulate the main result of the paper, showing when performance guarantees can be made. A detailed inverted pendulum example is presented in Section~\ref{sec:example} and 
conclusions are drawn in Section~\ref{sec:conclusion}.

\section{Related work}
\label{sec:related_work}
It is well-known that learning algorithms might cause unsafe behavior, both during training and possibly even after training,
as it can be hard to guarantee performance in all cases. Therefore, safety of learning approaches is a very active research area, \cite{saunders2018trial,chow2018lyapunov,el2016convex,perkins2002lyapunov,berkenkamp2017safe}.

In \cite{el2016convex} Constrained Markov Decision Problems (CMDPs) were used, and the cumulative cost was replaced by a stepwise one, which was then transferred into 
the admissible control set leading back to a standard MDP formulation.

There is also a set of approaches using Lyapunov ideas, originating in control theory.
In \cite{perkins2002lyapunov} a Lyapunov approach was used to guarantee stability of an RL agent.
The agent was allowed to switch between a given set of controllers that were designed to be safe
no matter what switching strategy was used. 
Then, in \cite{berkenkamp2017safe}, Lyapunov concepts were used to iteratively estimate the region of attraction of the policy, i.e., the region that the state 
is guaranteed not to leave, when applying the controller at hand. At the same time, while being in this safe region, the estimate of the region, as well as performance, was improved.
Finally, Chow et al. used the CMDPs to construct Lyapunov functions using linear programming, \cite{chow2018lyapunov}.
The approach is guaranteed to provide feasible, and under certain assumptions, optimal policies.

Our approach differs from 
\cite{saunders2018trial,chow2018lyapunov,el2016convex,perkins2002lyapunov,berkenkamp2017safe}
 by not trying to build the performance guarantees into the learning controller, but leveraging the reactivity of the surrounding BT structure and the existing model-based controller to create a combination with the required guarantees.

BTs were invented in the gaming industry \cite{champandard2010behavior} and are currently spreading throughout the fields of robotics and AI \cite{colledanchise2018book}. Significant effort to combine BTs with learning from experience as well as demonstrations can be found in the literature \cite{lim2010evolving,colledanchise2015learning,nicolau2016evolutionary,pereira2015framework,dey2013ql,fu2016reinforcement,french2019demonstration,sagredo2017trained,paxton2016costar}. 

In \cite{colledanchise2015learning}, a complete sub-tree is learned using a genetic algorithm applied to the Mario AI environment. Similar ideas were explored in \cite{lim2010evolving}.
Furthermore, grammatical evolution was used in  \cite{nicolau2016evolutionary}, to optimize the structure of a BT playing a platform game, while constraining the design to an and-or tree structure deemed efficient for the problem at hand.

Classical reinforcement learning was applied to BTs in 
\cite{pereira2015framework}, where the idea of replacing a given action (leaf) node with an RL policy was proposed.
Replacing non-leaf nodes with an RL policy deciding which child to execute was explored in 
\cite{dey2013ql}, and \cite{fu2016reinforcement}.

In \cite{french2019demonstration} the BT for performing pick and place operations were learned from human demonstration, using logic and decision trees.
A related idea was used in \cite{sagredo2017trained}, where a game designer first controlled game characters to create a database of trajectories that are then used to learn controllers.
Finally, in \cite{paxton2016costar}, a framework for end user instruction of a robot assistant was proposed.

Our approach differs from
\cite{colledanchise2015learning,sagredo2017trained,nicolau2016evolutionary,fu2016reinforcement,dey2013ql,pereira2015framework,paxton2016costar,french2019demonstration,lim2010evolving}
 by not focusing on how to integrate learning into a BT, but instead providing safety and performance guarantees when such learning has been integrated. Thus, the proposed approach can be combined with any of the methods described in \cite{colledanchise2015learning,sagredo2017trained,nicolau2016evolutionary,fu2016reinforcement,dey2013ql,pereira2015framework,paxton2016costar,french2019demonstration,lim2010evolving}.


\section{Background}
\label{sec:background}

In this section, we provide background on the formulation of BTs as discontinuous dynamical systems, following \cite{sprague2021continuous}.

\subsection{Continuous-time behavior trees}

We will now provide the continuous-time formulation of BTs, following \cite{sprague2021continuous}.
    Throughout this paper, let $\mathbb{R}^n$ be the state space, $x \in \mathbb{R}^n$ be a state, $\mathbb{R}^m$ be the control space, and $u \in \mathbb{R}^m$ be a control input.

\begin{definition}[Behavior Tree]\label{def:behavior}
    A function 
     $\bt_i: \mathbb{R}^n \to \mathbb{R}^m \times \{\mathcal{R},\mathcal{S},\mathcal{F}\}$,
     defined as
    \begin{equation}\label{eq:behavior}
        \bt_i(x):=\left(u_i\left(x\right),r_i\left(x\right) \right),
    \end{equation}
    where $x \in \mathbb{R}^n$ is the state, $i \in V$ is an index,
    $u_i: \mathbb{R}^n \rightarrow  \mathbb{R}^m$ is a controller,
    and $r_i: \mathbb{R}^n \rightarrow  \{\mathcal{R},\mathcal{S},\mathcal{F}\}$ 
    is a metadata function, describing the progress of the controller in terms of the 
    outputs:
    running ($\mathcal{R}$),
    success ($\mathcal{S}$), 
    and
    failure ($\mathcal{F}$).
    Define the metadata regions for $x \in \mathbb{R}^n$ as the running, success, and failure regions:
    \begin{equation}\label{eq:metadata-regions}
\begin{gathered}
    R_i :=\left\{x: r_i(x)=\mathcal{R} \right\}, \\
    S_i :=\left\{x: r_i(x)=\mathcal{S} \right\}, \quad F_i :=\left\{x: r_i(x)=\mathcal{F} \right\},
\end{gathered}
    \end{equation}
    respectively,
    which are pairwise disjoint and cover $\mathbb{R}^n$.
\end{definition}

The intuition of the metatdata regions (\ref{eq:metadata-regions}) is as follows.
If $x \in S_i$, then $\bt_i$ has either succeeded at achieving its goal
(e.g. opening a door) or the goal was already achieved (the door was already open).
Either way, it might make sense to execute another BT in sequence to achieve another goal (perhaps a goal that is intended to be achieved after opening the door).

If $x \in F_i$, then $\bt_i$ has either failed (the door to be opened turned out to be locked), or has no chance at succeeding (the door is impossible to get to from the current position).
Either way, it might make sense to execute another BT as a fallback (either to open the door in another way or to achieve a higher-level goal in a way that does not involve opening the door).

If $x \in R_i$, then it is too early to determine if $\bt_i$ will succeed or fail.
In most cases, it makes sense to continue executing $\bt_i$, but it could also be reasonable execute another BT if some other goal is more important (e.g. low battery levels indicate the need to recharge).

Above, the term ``execution'' means the use of a BT's controller $u_i$ in some underlying system.
Such a system can be seen as a dynamical system,
thus we have the following definition of a BT execution.

\begin{definition}[BT execution]\label{def:execution}
    A dynamical system given for a BT $\bt_i$ as
    \begin{equation}
        \label{eq:execution}
        \dot{x} = f\left(x,u_i\left(x\right)\right),
    \end{equation}
    where $f : \mathbb{R}^n \times \mathbb{R}^m \to \mathbb{R}^n$ is a system to be controlled
    and $u_i$ is given by (\ref{eq:behavior}).
\end{definition}

As shown in \cite{sprague2021continuous}, the BT execution (\ref{eq:execution})
    can be characterized by a \textit{discontinuous dynamical system} (DDS) defined over a finite set of so-called \textit{operating regions} \cite[Theorem 2, p.5]{sprague2021continuous}, with corresponding results regarding existence and uniqueness of its solutions \cite[Theorem 3, p.6]{sprague2021continuous}.
    These operating regions (defined below) arise from the switching among BTs invoked by \textit{BT compositions}, of which there exists two fundamental types: \textit{Sequences} and \textit{Fallbacks}, which we will now define.

    A Sequence is a BT that composes together sub-BTs that are to be executed in sequence, where each one requires the success of the previous.
    It will succeed only if all sub-BTs succeed, whereas, if any one sub-BT runs or fails, it will run or fail.
    This behavior is formalized in terms of the sub-BT's metadata regions in the following definition.

\begin{definition}[Sequence]
    A function
    $Seq$ 
    that composes an arbitrarily finite sequence of 
    $M \in \mathbb{N}$  BTs
    into a new BT as
    \begin{equation}\label{eq:sequence}
        Seq\left[\bt_1, \dots, \bt_M\right]\left(x\right) :=
        \begin{cases}
        \bt_M\left(x\right) & \text{if} \quad x \in S_1 \cap \dots S_{M-1} \\
        \vdots & \vdots \\
        \bt_2\left(x\right) & \text{else-if} \quad  x \in S_1 \\
        \bt_1\left(x\right) & \text{else}.
        \end{cases}
    \end{equation}
\end{definition}

A Fallback, on the other hand, is a BT that composes together sub-BTs that are to be executed as a fallback to one another, where each one is executed only in case of failure of the previous.
It will fail only if all sub-BTs fail, whereas, if any one sub-BT runs or succeeds, it will run or succeed.
This behavior is formalized in terms of the sub-BT's metadata regions in the following definition.

\begin{definition}[Fallback]
    A function
    $Fal$ 
    that composes an arbitrarily finite sequence of 
    $M \in \mathbb{N}$  BTs
    into a new BT as
    \begin{equation}\label{eq:fallback}
        Fal\left[\bt_1, \dots, \bt_M\right]\left(x\right) :=
        \begin{cases}
        \bt_M\left(x\right) & \text{if} \quad x \in F_1 \cap \dots F_{M-1} \\
        \vdots & \vdots \\
        \bt_2\left(x\right) & \text{else-if} \quad  x \in F_1 \\
        \bt_1\left(x\right) & \text{else}.
        \end{cases}
    \end{equation}
\end{definition}

\subsection{BTs as discontinuous dynamical systems}

As mentioned above, the execution (\ref{eq:execution}) can be characterized by a DDS defined over so-called \textit{operating regions}.
The state's presence in these operating regions can be viewed as the sufficient conditions for a sub-BT to be executed by the root BT $\bt_0$.
We will now formalize this in the following theorem from \cite{sprague2021continuous}.

\begin{theorem}[Operating regions]\label{theorem:dds}
    Assuming $\bt_0$ is the root BT, 
    there exists a maximum subset of the index set $P \subseteq V$ and an operating region $\Omega_i \subseteq \mathbb{R}^n$ for each index such that $\{\Omega_i\}_{i \in P}$ is a partition of $\mathbb{R}^n$ and
    \begin{equation}
        x \in \Omega_i \implies \bt_0\left(x\right) = \bt_i\left(x\right).
    \end{equation}
\end{theorem}
\begin{proof}
    See \cite[Theorem 2, p.5]{sprague2021continuous}
\end{proof}

The maximum subset $P \subseteq V$ in Theorem \ref{theorem:dds} corresponds to the set of leaf nodes
because it must be a set of nodes such that none of them are a parent to one another \cite{sprague2021continuous} --- the maximum set fulfilling this criteria is the leaf nodes.
Further, $P$ only includes such leaf nodes whose operating regions are not empty, because $\{\Omega_i\}_{i \in P}$ is a partition of $\mathbb{R}^n$.
For more detail on the derivation of the operating regions $\Omega_i$, see \cite{sprague2021continuous}.

The significance of Theorem \ref{theorem:dds} is that we can now interpret the execution (\ref{eq:execution}) of the root BT $\bt_0$ as a DDS, where $x \in \Omega_i$ implies $\dot{x} = f(x, u_0(x)) = f(x, u_i(x))$.
We will use this formalism in the remainder of the paper.

\section{Main result}\label{sec:main_result}

The general problem that we will address in this paper is as follows.

\begin{problem}\label{prob:0}
    Given a system $f: \mathbb{R}^n \times \mathbb{R}^m \to \mathbb{R}^n$ to be controlled
    with a running cost $L : \mathbb{R}^n \times \mathbb{R}^m \to \mathbb{R}$ such that last state is the accumulated running cost with $\dot{x}_n = L(x,u)$,
    and controllers $u_S, u_{MB}, u_{DD} : \mathbb{R}^n \to \mathbb{R}^m$,
    where $u_S$ is a safety controller, $u_{MB}$ is a model-based controller with formal guarantees,
    and $u_{DD}$ is a data-driven controller without formal guarantees (but potentially lower running cost than $u_{MB}$) ---
    how can $u_S$, $u_{MB}$, and $u_{ML}$ be composed together with a BT $\bt_0$ to create a controller $u_0$ that
    exploits $u_{MB}$, 
    yet still 
    formally guarantees that solutions to the execution (\ref{eq:execution}) of $\bt_0$
    avoids unsafe regions  $O \subset \mathbb{R}^n$
    and
    converges to a goal region $G \subset \mathbb{R}^n \setminus O$?
\end{problem}



\subsection{BT structure}

The BT $\bt_0$ we propose to solve Problem \ref{prob:0} is shown in Fig. \ref{fig:cyclical_bt}, and is formalized as follows.

\begin{assumption}[BT structure]\label{as:ex_bt}
    $\bt_0$ is given by
    \begin{equation}\label{eq:ex_bt}
        \bt_0 = Seq\left[\bt_1, \bt_2, \bt_5\right] \quad \text{s.t.} \quad \bt_2 = Fal\left[\bt_3, \bt_4\right]
    \end{equation}
    where 
    $\bt_1 = (u_S, r_S)$ is a safety action,
    $\bt_3 = (u_L, r_C)$ is a high-cost condition,
    $\bt_4 = (u_{DD}, r_G)$ is a data-driven action,
    $\bt_5 = (u_{MB}, r_G)$ is a model-based action,
    $r_S$ is a safety metadata function,
    $r_G$ is a goal metadata function,
    and
    $u_L$ is an arbitrary controller.
\end{assumption}

The BT structure in Assumption \ref{as:ex_bt} is intended to work as follows.
The data-driven action $\bt_4$ will be executed as long as it does not bring the system too close to the unsafe region or accumulate too much running cost.
If the system gets too close to the unsafe region then $\bt_1$ should execute until the system is sufficiently far away.
If too much running cost is accumulated then the model-based action $\bt_5$ should execute instead of the data-driven one.
If the data-driven action brings the system close enough to the goal region, then the model-based action should takeover the execution, as it will be formally guaranteed to stay in the goal region with negligible cost.
We formalize the above in the following assumption on the metadata regions formed by the metadata functions: $r_1 = r_S$, $r_3 = r_C$, and $r_4 = r_5 = r_G$.

\begin{assumption}[Metadata regions]\label{as:ex_overlaps}
    The metadata regions are chosen such that the following holds,
    \begin{equation}\label{eq:ex_overlaps}
        \begin{aligned}[t]
            F_1 &= O, \\
        \end{aligned}
        \qquad
        \begin{aligned}[t]
            R_3 &= \emptyset, \\
            S_3 &= C,
        \end{aligned}
        \qquad
        \begin{aligned}[t]
            S_4 &= S_5 = G, \\
            F_4 &= F_5 = \emptyset.
        \end{aligned}
    \end{equation}
    
    That is, there exists an unsafe region $O \subset \mathbb{R}^n$ that the failure region of the safety action $\bt_1$ is equal to,
    the running region of the high-cost condition $\bt_3$ is empty (guaranteeing that the arbitrary $u_L$ never executes),
    there exists a high-cost region $C \subset O^c$ that the success region of the high-cost condition $\bt_3$ is equal to,
    there exists a goal region $G \subset O^c$ that the success regions of the data-driven action $\bt_4$ and model-based action $\bt_5$ are equal to,
    and
    the failure regions of the data-driven action $\bt_4$ and model-based action $\bt_5$ are empty (they can be executed from any state).
    
\end{assumption}

We will now identify the operating regions of the BT in Fig. \ref{fig:cyclical_bt}, given by Assumptions \ref{as:ex_bt} and \ref{as:ex_overlaps}, following \cite{sprague2021continuous}.

\begin{lemma}
    If Assumptions \ref{as:ex_bt} and \ref{as:ex_overlaps} hold,
    the operating regions of Theorem \ref{theorem:dds} are given for $\bt_0$ as
    \begin{equation}\label{eq:ex_operating}
        \Omega_1 = S_1^c, \quad 
        \Omega_4 = S_1 \cap \left(C \cup G\right)^c, \quad
        \Omega_5 = S_1 \cap \left(C \cup G\right).
    \end{equation}
\end{lemma}
\begin{proof}

    Following \cite[Def. 5, p.4]{sprague2021continuous}, the so-called influence regions are
    \begin{equation}\label{eq:ex_influence}
        \begin{aligned}[t]
            I_0 &= I_1 = \mathbb{R}^n \\
            I_2 &= I_3 = S_1
        \end{aligned}
        \qquad
        \begin{aligned}[t]
            I_4 &= S_1 \cap F_3 \\
            I_5 &= S_1 \cap \left[S_3 \cup \left(F_3 \cap S_4\right)\right].
        \end{aligned}
    \end{equation}
    Following \cite[Def. 6, p.5]{sprague2021continuous}, the so-called success and failure pathways are
    $\mathfrak{S} = \left\{0, 5\right\}$ and $\mathfrak{F} = \left\{0, 1, 2, 4, 5 \right\}$
    respectively.
    Following \cite[Def. 7, p.5]{sprague2021continuous},
    the operating regions are
    \begin{equation}\label{eq:ex_pre_operating}
        \begin{aligned}
            \Omega_0 &= \Omega_1 \cup \Omega_2 \cup \Omega_5 \\
            \Omega_1 &= R_1 \cup F_1 \\
            \Omega_2 &= \Omega_3 \cup \Omega_4 \\
        \end{aligned}
        \qquad
        \begin{aligned}
            \Omega_3 &= S_1 \cap R_3 \\
            \Omega_4 &= S_1 \cap F_3 \cap \left(R_4 \cup F_4\right) \\
            \Omega_5 &= S_1 \cap \left[S_3 \cup \left(F_3 \cap S_4\right)\right].
        \end{aligned}
    \end{equation}

    Metadata regions are pairwise disjoint and cover $\mathbb{R}^n$, thus, by (\ref{eq:ex_overlaps}), we have that $F_1 = O$ implies $R_1 = (S_1 \cup O)^c$,
    $R_3 = \emptyset$ and $S_3 = C$ imply $F_3 = C^c$,
    and
    $S_4 = S_5 = G$ and $F_4 = F_5 = \emptyset$ imply $R_4 = R_5 = G^c$.
    Thus, with the full application of (\ref{eq:ex_overlaps}), including $R_3 = \emptyset$, on (\ref{eq:ex_pre_operating}), the non-empty operating regions of the leaf nodes are given in (\ref{eq:ex_operating}).
\end{proof}

The important thing to note in the operating regions (\ref{eq:ex_operating}) is that the high-cost region $C$ and the goal region $G$ are given.
Thus the success region $S_1$ of the safety action $\bt_1$ is the only region left to be specified.
In the next section we will show how the specification of $S_1$ affects the execution of $\bt_0$.


\subsection{Policy-level stability}

As discussed in the previous section, the specification of a BT's structure and its metadata regions determine its operating regions, which determine where a particular sub-BT will be executed.
However, in order to understand where a sub-BT's execution will lead to, we must look at the stability properties of its execution.
To do this, we first have the following definitions, in a similar spirit to \cite{colledanchise2017behavior}.

\begin{definition}[Finite-time successful]\label{def:fts}
    A BT $\bt_i$ is FTS if there exists a region $B_i \subseteq R_i \cup S_i$ that is positively invariant
    under its execution (\ref{eq:execution}),
    and a finite time $\tau_i \in [0, \infty) \subset \mathbb{R}$
    such that
    for all of its execution's solutions we have that
    $x(0) \in B_i$ implies $x(t) \in S_i \cap B_i$ in finite time $t \in [0, \tau_i]$.
\end{definition}


\begin{definition}[Safe]\label{def:safe}
    A BT $\bt_i$ is safe w.r.t. an unsafe region $O \subset \mathbb{R}^n$
    if there exists a region $B_i \subseteq \mathbb{R}^n \setminus O$ that is positively invariant
    under its execution (\ref{eq:execution}).
\end{definition}

\begin{definition}[Safeguarding]\label{def:safeguarding}
    A BT $\bt_i$ is safeguarding w.r.t. an unsafe region $O \subset \mathbb{R}^n$
    if it is FTS and safe with $S_i \subset B_i$.
\end{definition}

\say{Finite-time success} means that, for the execution of a sub-BT, if the state starts in a certain region $B_i$ then it will be in the success portion of this region $B_i \cap S_i$ in finite time, without every venturing out of it.
\say{Safe} means that, if the state starts in $B_i$ then it will never venture into the unsafe region $O$.
\say{Safeguarding} means that if the state exits the success region $S_i$, then it will be redirected back into it in finite time without venturing into the unsafe region, because the region $B_i$ surrounds it, $S_i \subset B_i$.
One way to guarantee the above properties is through \textit{exponential stability}, as we show in the following lemma, in a similar spirit to \cite{colledanchise2017behavior}.

\begin{lemma}[Exponential stability]\label{lemma:exponential_stability}
    A BT $\bt_i$ for which 
    there exists a locally exponentially stable equilibrium
    $x_i \in B_i \cap S_i$ 
    on a region $B_i \subseteq R_i \cup S_i$
    for its execution (\ref{eq:execution})
    is FTS.
    Additionally, if there exists an unsafe region $O \subset \mathbb{R}^n$ such that 
    $B_i \subseteq \mathbb{R}^n \setminus O$ then $\bt_i$ is safe w.r.t. $O$,
    and if $S_i \subset B_i$ also then $\bt_i$ is safeguarding w.r.t. $O$.
\end{lemma}
\begin{proof}
    If the execution (\ref{eq:execution}) of $\bt_i$ is locally exponentially stable about $x_i \in B_i \cap S_i$
    on $B_i \subseteq R_i \cup S_i$
    then there exists
    $\alpha_i, \beta_i \in (0, \infty]$
    such that, for its solutions, if $x(0) \in B_i$ then
    $\Vert x(t) - x_i \Vert \leq  \Vert x(0) - x_i \Vert \alpha_i e^{- \beta_i t}$
    for all 
    $t \in [0, \infty)$.
    The stability of $x_i \in B_i \cap S_i$ implies that there must exist
    a maximal 
    $\epsilon_i \in (0, \infty)$ 
    and minimal 
    $\tau_i \in (0, \infty)$
    such that
    $\Vert x(\tau_i) - x_i \Vert \leq  \Vert x(0) - x_i \Vert \alpha_i e^{- \beta_i \tau_i} < \epsilon_i$
    and $\{x \in \mathbb{R}^n : \Vert x - x_i \Vert \leq \epsilon_i\} \subseteq B_i \cap S_i$.
    Thus, we must have that if $x(0) \in B_i$ then $x(t) \in B_i \cap S_i$ in finite time $t \in [0, \tau_i]$,
    and that $x(t) \in B_i \cap S_i$ for all 
    $t \in (\tau_i, \infty)$.
    If additionally $B_i \subseteq \mathbb{R}^n \setminus O$ then $\bt_i$ is safe (Def. \ref{def:safe}),
    and if $S_i \subset B_i$ also then $\bt_i$ is safeguarding (Def. \ref{def:safeguarding}).
\end{proof}

The stability properties described above allow us to make formal guarantees on how a sub-BT's execution will behave.
But, what can be said about how a sub-BT's behavior will lead to the execution of other sub-BTs?

To answer this, we must analyze how the aforementioned region $B_i$ overlaps with other sub-BTs' operating regions.
Informally speaking, if a sub-BT $\bt_i$ is FTS and $B_i$ does not intersect any other sub-BTs' operating regions, then the state will stay in $\bt_i$'s operating region and converge to success.
However, if $B_i$ does intersect other sub-BTs' operating regions, then there is a possibility that the state could venture into those operating regions.
Thus, with a strategic choice of $B_i$ for each sub-BT, an ordered set of possible sub-BT executions can be derived in a similar spirit to the sequential composition of 
\cite{burridge1999sequential, conner2003composition}.
We formalize this in the following theorem.

\begin{theorem}[Sequential composition]\label{theorem:sequential_composition}
    If a sub-BT $\bt_i$ is FTS,
    there exists a minimum subset $P_i \subset P \setminus \{i\}$, where $P$ is from Theorem \ref{theorem:dds},
    such that
    \begin{equation}\label{eq:sequential_composition}
        \Omega_i \cap B_i \neq \emptyset \qquad 
        B_i \setminus \Omega_i \subseteq \bigcup_{j \in P_i} \Omega_j \setminus F_j,
    \end{equation}
    and, for solutions to its execution (\ref{eq:execution}), 
    if $x(0) \in \Omega_i \cap B_i$, 
    then 
    either 
    $x(t) \in \Omega_i \cap S_i$
    or 
    $x(t) \in \Omega_j \cap (R_j \cup S_j)$ for some $j \in P_i$ 
    in finite-time $t \in [0, \tau_i]$.
\end{theorem}
\begin{proof}


    If $P_i = \emptyset$, then we have that $B_i \setminus \Omega_i \subseteq \emptyset$, which implies that $B_i \subseteq \Omega_i$.
    Since $\bt_i$ is FTS, it is implied that $B_i \cap S_i \neq \emptyset$.
    Thus, for the execution (\ref{eq:execution}) of $\bt_0$, if $x(0) \in \Omega_i \cap B_i$ then $x(t) \in \Omega_i \cap B_i \cap S_i$ in finite time $t \in [0, \tau_i]$.

    If $L_i \neq \emptyset$, 
    then $\{\Omega_j \setminus F_j\}_{j \in L_i}$ is a partition of $B_i \setminus \Omega_i$ because $L_i$ is minimum.
    Thus, for the execution (\ref{eq:execution}) of $\bt_0$, 
    if $x(0) \in \Omega_i \cap B_i$ then either 
    $x(t) \in \Omega_i \cap B_i \cap S_i$ or
    $x(t) \in \Omega_j \cap (R_j \cup S_j)$ in finite time $t \in [0, \tau_i]$.
\end{proof}

\subsection{Convergence}

We will now make an assumption on the stability properties of the BT proposed in Assumptions \ref{as:ex_bt} and \ref{as:ex_overlaps}, and then use the concept of Theorem \ref{theorem:sequential_composition} to prove convergence.

The assumption is motivated by the following.
If we are executing the data-driven sub-BT $\bt_4$, we want to ensure that it does not execute forever if the state does not get into the goal region $G$ or close to the unsafe region $O$.
Thus, we assume that the accumulated running cost will reach a finite value in finite time.
Further, we assume that the safety sub-BT $\bt_1$ is safeguarding on the region outside of the unsafe area, i.e. $B_1 = O^c$, so that if the data-driven sub-BT $\bt_4$ misbehaves, the system will not go into the unsafe region $O$.
Lastly, we assume that the model-based sub-BT $\bt_5$ is FTS.
We formalize this assumption as follows.

\begin{assumption}[Stability]\label{as:ex_cost_timeout}
    There exists a finite cost $J \in [0, \infty)$ such that
    the high cost region is given by $C = \mathbb{R}^{n-1} \times [J, \infty)$ and we have that $C \cap G \neq \emptyset$, and 
    a finite time $T \in [0, \infty)$
    such that if the state is not in the goal region before that time,
    $x(t) \not \in G$ for all $t \in [0, T]$,
    then the state will stay in the high cost region after that time,
    $x(t) \in C$ for all $t \in [T, \infty)$.
    The safety action is safeguarding w.r.t. $O$ such that $B_1 = O^c$,
    the the model-based action $\bt_5$ is FTS.
\end{assumption}

Now given Assumption \ref{as:ex_cost_timeout} and Theorem \ref{theorem:sequential_composition}, we will show how a strategic choice of the safety sub-BT's success region $S_1$ facilitates convergence from all starting states outside of the unsafe area.
In the following Theorem, we show how the overall BT design is convergent if
we set $S_1$ equal to the goal region and the region for which the model-based sub-BT is FTS.

\begin{theorem}[Constant metadata]\label{theorem:no_switch}
    If Assumptions 
    \ref{as:ex_bt}, 
    \ref{as:ex_overlaps}, and 
    \ref{as:ex_cost_timeout} hold,
    and we have that $S_1 = G \cup B_5$,
    then $\bt_0$ is FTS with $\tau_0 = T + \tau_1 + \tau_5$ 
    and safe w.r.t. $O$ starting from $x(0) \not \in O$.
\end{theorem}
\begin{proof}
    We have that if $x(0) \not \in O$ then either $x(0) \in R_1$ or $x(0) \in S_1$ because $F_1 = O$ and metadata regions are pairwise disjoint.

    If $x(0) \in R_1$ then $x(0) \in \Omega_1 \cap B_1$ because 
    $\Omega_1 = S_1^c = R_1 \cup O$ and $B_1 = O^c = R_1 \cup S_1$.
    We have that $B_1 \setminus \Omega_1 = S_1$ because $S_1 \setminus O = S_1$ since metadata regions are pairwise disjoint.
    We have that $\Omega_4 \setminus F_4 = \Omega_4$ because $F_4 = \emptyset$
    and
    $\Omega_5 \setminus F_5 = \Omega_5$
    because $F_5 = \emptyset$.
    We then have that $\Omega_4 \cup \Omega_5 = S_1 \cap ((C \cup G)^c \cup (C \cup G)) = S_1$.
    Thus we have that $B_1 \setminus \Omega_1 \subseteq (\Omega_4 \setminus F_5) \cup (\Omega_5 \setminus F_5)$. 
    Thus from Theorem \ref{theorem:sequential_composition}, we have that
    $x(0) \in \Omega_1 \cap B_1$ implies either 
    $x(t) \in \Omega_4 \cap (R_4 \cup S_4) = \Omega_4$ or 
    $x(t) \in \Omega_5 \cap (R_5 \cup S_5) = \Omega_5$,
    and hence $x(t) \in S_1$ 
    in finite time $t \in [0, \tau_1]$.

    If $x(0) \in S_1$ then either 
    $x(0) \in \Omega_4$ or 
    $x(0) \in \Omega_5$ because 
    $\Omega_4 \subseteq S_1$ and $\Omega_5 \subseteq S_1$.

    If $x(0) \in \Omega_4$, then either 
    $x(t) \in \Omega_5$ or 
    $x(t) \in \Omega_1$ (in which case $x(t) \in \Omega_1 \cap B_1$ because $B_1 \supset S_1$ since $\bt_1$ is safeguarding)
    in finite time $t \in [0, T]$,
    where $T \in [0, \infty)$ is a finite time for which either $x(t) \in G$ or $x(t) \in C$ by Assumption \ref{as:ex_cost_timeout}.

    If $x(0) \in \Omega_5$ then $x(t) \in \Omega_5 \cap G$ in finite time $t \in [0, \tau_5]$ for the following reasons.
    Since $S_1 = G \cup B_5 $, we have that $\Omega_5 = G \cup (B_5 \cap C)$,
    and since $C$ is positively invariant under the execution (\ref{eq:execution}) of $\bt_0$ by Assumption \ref{as:ex_cost_timeout},
    we have that $B = B_5 \cap C \neq \emptyset$ is also positively invariant under the execution.
    We have that $\Omega_5 \cap B \neq \emptyset$ and
    $B \setminus \Omega_5 = (B_5 \cap C) \cap G^c \cap (B_5 \cap C)^c = \emptyset$.
    Thus, by Lemma \ref{theorem:sequential_composition},
    we have that if $x(0) \in \Omega_5 \cap B_5$ then $x(t) \in \Omega_5 \cap B_5 \cap G$ in finite time $t \in [0, \tau_i]$.

    Thus, for the execution (\ref{eq:execution}) of $\bt_0$, we have that
    if $x(0) \not \in O$ then $x(t) \in G$ in finite time $t \in [0, \tau_0]$,
    where $\tau_0 = T + \tau_1 + \tau_5$.
    Thus, $B_0 = B_1$ is a region for which $\bt_0$ is FTS and safe.
\end{proof}

One limitation of Theorem \ref{theorem:no_switch} is that $S_1 = G \cup B_5$ might be hard to achieve and verify.
One way to loosen this requirement is by switching the success region of the safety sub-BT, thereby affecting where the sub-BTs will be executed according to (\ref{eq:ex_operating}).

If, at first, we have $S_1 \subset G \cup B_5$, where $B_5$ is the region with which the model-based sub-BT is FTS, then the safety sub-BT will bring the state to that region in finite time.
If we then switch the success region so that $S_1 \supset G \cup B_5$, then either data-driven sub-BT $\bt_4$ or the model-based sub-BT $\bt_5$ will be executed.
If $\bt_5$ is executed, then it will successfully bring the state to the goal region $G$ because the region $B_5$ for which it is FTS is contained in $S_1$.
However, if $\bt_4$ is executed and it misbehaves so that the state ventures out of $S_1 \supset G \cup B_5$, then we can switch to $S_1 \subset G \cup B_5$ again, bring the state back again with $\bt_1$, switch back to $S_1 \supset G \cup B_5$, and so on.
This behavior is formalized in the following Theorem, in a similar spirit to Theorem \ref{theorem:no_switch}.

\begin{theorem}[Switching metadata]\label{theorem:switch}
    If Assumptions 
    \ref{as:ex_bt}, 
    \ref{as:ex_overlaps}, and 
    \ref{as:ex_cost_timeout} hold,
    and we have that
    \begin{equation}\label{eq:success_switch}
        S_1 \gets \begin{cases}
            S_1' &\text{if} \quad x \not \in S_1'' \\
            S_1'' &\text{if} \quad x \in S_1'
        \end{cases}
        \quad \text{s.t.} \quad
        \begin{aligned}
            S_1' &\subset G \cup B_5 \\
            S_1'' &\supset G \cup B_5,
        \end{aligned}
    \end{equation}
    then $\bt_0$ is FTS with $\tau_0 = T + \tau_1 + \tau_5$ and safe w.r.t. $O$ starting from $x(0) \in O$.
\end{theorem}
\begin{proof}
    If $x(0) \not \in O$ then either
    $x(0) \not \in S_1''$,
    $x(0) \in S_1''$,
    or $x(0) \in S_1$,
    because $S_1' \subset G \cup B_5 \subset S_1''$ and $F = O$.

    If $x(0) \not \in S_1''$ then $S_1 \gets S_1'$ by (\ref{eq:success_switch}),
    in which case $x(0) \not \in S_1$ and hence $x(0) \in R_1$.
    In the same way as Theorem \ref{theorem:no_switch}, 
    if $x(0) \in R_1$ then $x(t) \in S_1$ in finite time $t \in [0, \tau_1]$.

    If $x(0) \in S_1$ and $S_1 = S_1'$ then $S_1 \gets S_1''$ and either 
    $x(0) \in \Omega_4$ or 
    $x(0) \in \Omega_5$ because 
    $\Omega_4 \subseteq S_1$ and 
    $\Omega_5 \subseteq S_1$.

    If $x(0) \in \Omega_4$,
    in the same way as Theorem \ref{theorem:no_switch},
    if $x(0) \in \Omega_4$, then either 
    $x(t) \in \Omega_5$ or 
    $x(t) \in \Omega_1$ (in which case $x(t) \in \Omega_1 \cap B_1$ because $B_1 \supset S_1$ since $\bt_1$ is safeguarding)
    in finite time $t \in [0, T]$,
    where $T \in [0, \infty)$ is a finite time for which either $x(t) \in G$ or $x(t) \in C$ by Assumption \ref{as:ex_cost_timeout}.

    If $x(0) \in \Omega_1$ and $S_1 = S''$ then $S_1 \gets S_1'$ because 
    $x(0) \not \in S''$.
    Thus, again, $x(t) \in S_1$ with $S_1 = S'$ in finite time $t \in [0, \tau_1)$.

    If $x(0) \in \Omega_5$ and $S_1 = S_1'$ then
    $S_1 \gets S_1''$ and $x(0) \in G \cup B_5$ because
    $S_1' \subset G \cup B_5$.
    We have that $B = B_5 \cap C$ is positively invariant under the execution (\ref{eq:execution}) of $\bt_5$.
    We have that $B \setminus \Omega_5 = \emptyset$ because now $S_1 \supset G \cup B_5$.
    Thus, by Theorem \ref{theorem:sequential_composition},
    we have that if $x(0) \in \Omega_5 \cap B_5$ with $S_1 = S_1''$ then $x(t) \in \Omega_5 \cap B_5 \cap G$ in finite time $t \in [0, \tau_i]$.

    Thus, for the execution (\ref{eq:execution}) of $\bt_0$, we have that
    if $x(0) \not \in O$ then $x(t) \in G$ in finite time $t \in [0, \tau_0]$,
    where $\tau_0 = T + \tau_1 + \tau_5$.
    Thus, $B_0 = B_1$ is a region for which $\bt_0$ is FTS and safe.
\end{proof}

\section{Example: Inverted pendulum}
\label{sec:example}
To illustrate the results above,
we will now assume that the given dynamical system to be controlled 
is given by the inverted pendulum of \cite{SRINIVASAN2002133}, with the state augmented by the cost function given in \cite{sprague2019learning}.

\begin{implementation}[Inverted pendulum]\label{imp:pendulum}
    The given system $f : \mathbb{R}^n \times \mathbb{R}^n \to \mathbb{R}^n$
    of Problem \ref{prob:0} is given by
    \begin{equation}\label{eq:pendulum}
        f\left(x, u\right)
        =
        \begin{bmatrix}
            v \\
            u \\
            \omega \\
            \sin \left(\theta\right) - u \cos\left(\theta\right) \\
            L(x, u)
        \end{bmatrix},
    \end{equation}
    where $x = [y, v, \theta, \omega, J] \in \mathbb{R}^5$ is the state,
    $u \in [-u_m, u_m]$ is a control input;
    $y$ and $v$ are the translational position and velocity of the cart;
    $\theta$ is the clockwise angle of the pole from the upright orientation and $\omega$ is the angular velocity of the pole;
    and $J$ is accumulated running cost,
    and $L$ is given by
    \begin{equation}\label{eq:lagrangian}
        L\left(x, u\right) = \alpha + \left(1 - \alpha\right) u^2 \quad \text{s.t.} \quad \alpha \in [0, 1] \subset \mathbb{R},
    \end{equation}
    where $\alpha$ is a chosen parameter.
\end{implementation}

The parameter $\alpha$ for the running cost function (\ref{eq:lagrangian}) will be used for both training the data-driven sub-BT and for satisfying Assumption \ref{as:ex_cost_timeout}, as we will describe in the next section.
If $\alpha = 0$, then (\ref{eq:lagrangian}) is a quadratic running cost on the control input; if $\alpha = 1$ then it is a running cost on time.

\subsection{Controllers}

In this section we will define the controllers for Problem \ref{prob:0} under Assumption \ref{as:ex_bt}.

The model-based controller $u_{mb}$ we will use is the globally asymptotically stable one of Srinivasan et al. \cite[Theorem 2, p.4]{SRINIVASAN2002133}.
With the inverted pendulum (\ref{eq:pendulum}) under the influence of this controller,
there exists a locally exponentially stable region about the stationary upright configuration 
$x \in \{0\}^2 \times \{\theta : \cos(\theta) = 1\} \times \{0\} \times \mathbb{R}$
and a globally defined finite time at which the state will enter this region.
Based on the above, we make the following assumption on 
the model-based sub-BT.

\begin{implementation}[Model-based BT]\label{imp:model_based}
    The model-based controller $u_5 = u_{MB}$ is given by the controller 
    of Srinivasan et al. \cite[Theorem 2, p.4]{SRINIVASAN2002133}.
    The goal region $G = S_5 \supset \{0\}^2 \times \{\theta : \cos(\theta) = 1\} \times \{0\} \times \mathbb{R}$ is the region in which the stationary upright configuration is locally exponentially stable for the execution (\ref{eq:execution}) of $\bt_5$ with Implementation \ref{imp:pendulum}.
\end{implementation}


In reality, the track length of the pendulum (\ref{eq:pendulum}) would not be infinitely long, but bounded.
Bounds on cart position and velocity for the inverted pendulum under the control of $u_{MB}$ are given in \cite{SRINIVASAN2002133} as
$|y| \leq  y_m = \pi^2 (\pi + u_m)$ and
$|v| \leq v_m = \pi u_m $
when starting from the stationary-downright configuration
$x(0) \in \{0\}^2 \times \{\theta : \cos(\theta) = -1\} \times \{0\} \times \mathbb{R}$.
Thus, we would need the track length to be at least $2 y_m$ in width.
Given this, what width within these boundaries would we need to set as safe, to ensure that the data-driven controller does not bring the system to a state where violating the track length is inevitable, i.e., going so fast that braking in time is impossible?

To answer this, suppose we start at the origin with $y(t_0) = 0$ and $v(t_0) = 0$,
apply the maximum control input $u_m$ until some position $y(t_1)$ and velocity $v(t_1)$, and thereafter apply the minimum control input $-u_m$
until the cart is brought to a halt at the upper bound of the track length $y(t_2) = y_m$ with $v(t_2) = 0$.
Since the cart is a double integrator, basic kinematics tells us that
$v(t_1)^2 = v(t_0)^2 + 2 u_m(y(t_1) - y(t_0))$ and
$v(t_2)^2 = v(t_1)^2 - 2 u_m(y(t_2) - y(t_1))$,
which gives us $y(t_1) = y_m/2$ and $v(t_1) = \sqrt{u y_m}$.

We want to ensure that, if the data-driven sub-BT $\bt_5$ is being used (when $x \in \Omega_5 = S_1 \cap (C \cup G)^c$), the safety sub-BT $\bt_1$ will be activated if the state ventures outside of these bounds.
However, we also want to ensure that, if the model-based sub-BT $\bt_5$ is being used (when $x \in \Omega_4 = S_1 \cap (G \cup C)$), the safety sub-BT $\bt_1$ will not activate when the state ventures outside of these bounds.
Thus we have the following assumption on the safety sub-BT.

\begin{implementation}[Safety BT]\label{imp:safety}
    The safety metadata region is given by $S_1$ in (\ref{eq:success_switch}) with
    \begin{equation*}
        \begin{aligned}
            S'_1 = &\left\{0\right\}^2 \times \left\{\theta : \cos(\theta) = -1\right\} \times \left\{0\right\} \times \mathbb{R} \\
            S''_1 = &\left(\left[-y_m, y_m\right] \times \left[-v_m, v_m\right] \times \mathbb{R}^3 \cap \left(G \cup C\right)\right) \\
            &\cup \left(\left[-\frac{y_m}{2}, \frac{y_m}{2}\right] \times \left[-\sqrt{u y_m}, \sqrt{u y_m}\right] \times \mathbb{R}^3 \cap \left(G \cup C\right)^c\right) \\
            &\text{s.t.} \quad y_m = \pi^2(\pi + u_m), \quad v_m = \pi u_m.
        \end{aligned}
    \end{equation*}
    The safety controller is given by
    \begin{equation*}
        u_S(x) = \begin{cases}
            -k_y y - k_v v &\text{if} \quad x \in S''_1 \\
            -u_m \text{sgn}\left(y\right) &\text{else}.
        \end{cases}
    \end{equation*}
    The obstacle region is given by
    \begin{equation*}
        O = [-y_m, y_m]^c \times [-\sqrt{u_m y_m}, \sqrt{u_m y_m}]^c \times 
        \mathbb{R}^3.
    \end{equation*}
\end{implementation}

Finally, we implement the data-driven sub-BT 
$\bt_4 = (u_{DD}, r_G)$ with the same goal metadata function as in Implementation \ref{imp:model_based} and a mulitlayer-perceptron (MLP) controller.

\begin{figure}[ht]
    \centering
    \includegraphics[width=\linewidth]{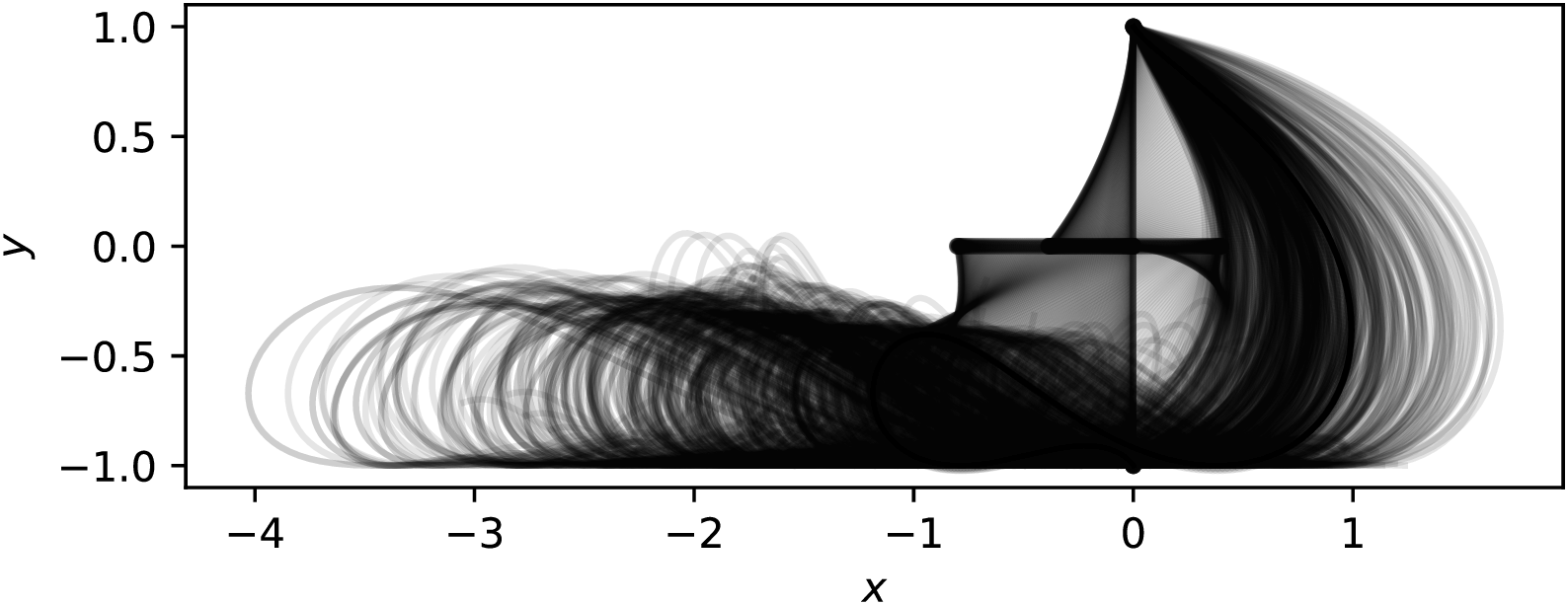}
    \caption{The database of optimal swingup trajectories used to train the data-driven controller $u_{DD}$, where $x$ and $y$ are horizontal and vertical coordinates, respectively.}
    \label{fig:pendulum}
\end{figure}

\begin{implementation}[Data-driven BT]\label{imp:nn}
    The data-driven controller $u_{DD}$ is a MLP with $3$ hidden layers, each having $50$ nodes, where the softplus activation function is used.
    The output layer is a $tanh$ activation scaled to $[-u_m, u_m]$.
    We train $u_{DD}$ for $354918$ iterations at a learning rate of $0.001$ with behavioral cloning
    on $354918$ state-control pairs (see Fig. \ref{fig:pendulum}) coming from optimal trajectories w.r.t (\ref{eq:lagrangian}) with $\alpha = 0$, which are obtained from Pontryagin's maximum principle in the way of \cite{sprague2019learning}.
    On $90\%$ of the data, $u_{DD}$ achieved $2.1445e^{-3}$ mean-squared error (MSE) loss in training, and $2.2745e^{-3}$ MSE loss in on the remaining data in testing.
\end{implementation}

Since the data-driven controller $u_{MB}$ is trained from trajectories that are optimal w.r.t. a quadratic cost on the control, i.e. (\ref{eq:lagrangian}) with $\alpha = 0$, it tries to conserve effort.
We want to use this controller as long as it does not take too long, so
we consider the high-cost region $C$ to be where the cost (\ref{eq:lagrangian}) with $\alpha = 1$ reaches a certain value.

\begin{theorem}
    Theorem \ref{theorem:switch} is satisfied with
    Assumptions \ref{as:ex_bt} and \ref{as:ex_overlaps},
    using Implementations \ref{imp:pendulum}, \ref{imp:model_based}, \ref{imp:safety}, and \ref{imp:nn}
    with $\alpha = 1$
    and $C = \mathbb{R}^4 \times [T, \infty)$,
    where $T \in [0, \infty)$ is a finite time.
\end{theorem}
\begin{proof}
    We need to show that Assumption \ref{as:ex_cost_timeout} is fulfilled.
    First, the accumulated cost $J$ is monotonically increasing when $\alpha =1$ in (\ref{eq:lagrangian}).
    Thus, for solutions of (\ref{eq:execution}) of $\bt_0$,
    we have
    $x(t) \in C$ for all $t \in [T, \infty)$ from all $x(0) \in \mathbb{R}^n$ if $x(t) \not \in G$ for $t \in [0, T)$.
    Second, the safety controller $u_S$ is a standard PD controller,
    thus there exists a positively invariant set $B_1 = O^c$
    upon which 
    $x_1 \in S'_1$
    is a locally exponentially stable equilibrium, 
    and we have $S_1 \subset B_1$, thus by Lemma \ref{lemma:exponential_stability}, 
    $\bt_1$ is safeguarding and there exists a finite time $\tau_1$ for which it is FTS.
    Third, by \cite{SRINIVASAN2002133}, 
    there exists a finite time $\tau_5$ for which the execution (\ref{eq:execution}) of $\bt_5$ enters the goal region $G$, thus $\bt_5$ is FTS.
    Thus, Theorem \ref{theorem:switch} is satisfied.
\end{proof}


\section{Conclusions}
\label{sec:conclusion}
Previous works have separately explored different ways of adding learning components to a BT, and approaches for building safety guarantees into learning controllers.

In this paper we have shown how the reactivity and modularity of BTs
enable a design where safety and convergence guarantees are provided on top of any learning controller. This was done using a natural combination of the learning controller with a safety controller and a convergent model based controller. 
Such a design might however introduce deadlocks where the different controllers work against each other.
The paper presents a set of conditions that guarantee that the proposed design does not suffer from such problems, and instead achieves both safety and convergence to the desired goal states.
An inverted pendulum example was used to illustrate the approach.


\section*{Acknowledgement}
This  work  was  supported  by  Stiftelsen  for StrategiskForskning  
(SSF)  through  the  Swedish  Maritime Robotics Centre (SMaRC) (IRC15-0046).

\bibliographystyle{IEEEtran}
\bibliography{cdc}

\begin{thebibliography}{10}
\providecommand{\url}[1]{#1}
\csname url@rmstyle\endcsname
\providecommand{\newblock}{\relax}
\providecommand{\bibinfo}[2]{#2}
\providecommand\BIBentrySTDinterwordspacing{\spaceskip=0pt\relax}
\providecommand\BIBentryALTinterwordstretchfactor{4}
\providecommand\BIBentryALTinterwordspacing{\spaceskip=\fontdimen2\font plus
\BIBentryALTinterwordstretchfactor\fontdimen3\font minus
  \fontdimen4\font\relax}
\providecommand\BIBforeignlanguage[2]{{%
\expandafter\ifx\csname l@#1\endcsname\relax
\typeout{** WARNING: IEEEtran.bst: No hyphenation pattern has been}%
\typeout{** loaded for the language `#1'. Using the pattern for}%
\typeout{** the default language instead.}%
\else
\language=\csname l@#1\endcsname
\fi
#2}}

\bibitem{champandard2010behavior}
A.~Champandard, M.~Dawe, and D.~Cerpa, ``Behavior trees: Three ways of
  cultivating strong ai,'' in \emph{Game Developers Conference, Audio Lecture},
  2010.

\bibitem{colledanchise2017behavior}
M.~Colledanchise and P.~{\"O}gren, ``{How Behavior Trees Modularize Hybrid
  Control Systems and Generalize Sequential Behavior Compositions, the
  Subsumption Architecture, and Decision Trees},'' \emph{IEEE Transactions on
  Robotics}, vol.~33, no.~2, pp. 372--389, 2017.

\bibitem{sprague2021continuous}
C.~I. Sprague and P.~{\"O}gren, ``Continuous-time behavior trees as
  discontinuous dynamical systems,'' \emph{IEEE Control Systems Letters},
  vol.~6, pp. 1891--1896, 2021.

\bibitem{iovino2020survey}
M.~Iovino, E.~Scukins, J.~Styrud, P.~{\"O}gren, and C.~Smith, ``A survey of
  behavior trees in robotics and ai,'' \emph{accepted for publication in
  Autonomous Robots (and on ArXiv)}, 2022.

\bibitem{Mnih2015}
\BIBentryALTinterwordspacing
V.~Mnih, K.~Kavukcuoglu, D.~Silver, A.~A. Rusu, J.~Veness, M.~G. Bellemare,
  A.~Graves, M.~A. Riedmiller, A.~Fidjeland, G.~Ostrovski, S.~Petersen,
  C.~Beattie, A.~Sadik, I.~Antonoglou, H.~King, D.~Kumaran, D.~Wierstra,
  S.~Legg, and D.~Hassabis, ``Human-level control through deep reinforcement
  learning,'' \emph{Nature}, vol. 518, no. 7540, pp. 529--533, 2015, {Accessed
  on: Apr 20, 2017}. [Online]. Available:
  \url{http://dx.doi.org/10.1038/nature14236}
\BIBentrySTDinterwordspacing

\bibitem{silver2016mastering}
D.~Silver, A.~Huang, C.~J. Maddison, A.~Guez, L.~Sifre, G.~Van Den~Driessche,
  J.~Schrittwieser, I.~Antonoglou, V.~Panneershelvam, M.~Lanctot,
  \emph{et~al.}, ``Mastering the game of go with deep neural networks and tree
  search,'' \emph{nature}, vol. 529, no. 7587, p. 484, 2016.

\bibitem{justesen2019deep}
N.~Justesen, P.~Bontrager, J.~Togelius, and S.~Risi, ``Deep learning for video
  game playing,'' \emph{IEEE Transactions on Games}, 2019.

\bibitem{biggar2020modularity}
O.~Biggar, M.~Zamani, and I.~Shames, ``On modularity in reactive control
  architectures, with an application to formal verification,'' \emph{to appear
  in ACM Transactions on Cyber-Physical Systems}, 2022.

\bibitem{saunders2018trial}
W.~Saunders, G.~Sastry, A.~Stuhlmueller, and O.~Evans, ``Trial without error:
  Towards safe reinforcement learning via human intervention,'' in
  \emph{Proceedings of the 17th International Conference on Autonomous Agents
  and MultiAgent Systems}.\hskip 1em plus 0.5em minus 0.4em\relax International
  Foundation for Autonomous Agents and Multiagent Systems, 2018, pp.
  2067--2069.

\bibitem{chow2018lyapunov}
Y.~Chow, O.~Nachum, E.~Duenez-Guzman, and M.~Ghavamzadeh, ``A lyapunov-based
  approach to safe reinforcement learning,'' \emph{arXiv preprint
  arXiv:1805.07708}, 2018.

\bibitem{el2016convex}
M.~El~Chamie, Y.~Yu, and B.~A{\c{c}}{\i}kmese, ``Convex synthesis of randomized
  policies for controlled markov chains with density safety upper bound
  constraints,'' in \emph{American Control Conference}, 2016, pp. 6290--6295.

\bibitem{perkins2002lyapunov}
T.~J. Perkins and A.~G. Barto, ``Lyapunov design for safe reinforcement
  learning,'' \emph{Journal of Machine Learning Research}, vol.~3, no. Dec, pp.
  803--832, 2002.

\bibitem{berkenkamp2017safe}
F.~Berkenkamp, M.~Turchetta, A.~Schoellig, and A.~Krause, ``Safe model-based
  reinforcement learning with stability guarantees,'' in \emph{Advances in
  Neural Information Processing Systems}, 2017, pp. 908--918.

\bibitem{colledanchise2018book}
M.~Colledanchise and P.~\"Ogren, \emph{Behavior Trees in Robotics and AI, an
  Introduction}.\hskip 1em plus 0.5em minus 0.4em\relax Chapman and Hall/CRC,
  2018.

\bibitem{lim2010evolving}
C.~Lim, R.~Baumgarten, and S.~Colton, ``{E}volving {B}ehaviour {T}rees for the
  {C}ommercial {G}ame {DEFCON},'' \emph{Applications of Evolutionary
  Computation}, pp. 100--110, 2010.

\bibitem{colledanchise2015learning}
M.~Colledanchise, R.~Parasuraman, and P.~{\"O}gren, ``{Learning of Behavior
  Trees for Autonomous Agents},'' \emph{IEEE Transactions on Games, DOI
  10.1109/TG.2018.2816806}, 2018.

\bibitem{nicolau2016evolutionary}
M.~Nicolau, D.~Perez-Liebana, M.~O'Neill, and A.~Brabazon, ``Evolutionary
  behavior tree approaches for navigating platform games,'' \emph{IEEE
  Transactions on Computational Intelligence and AI in Games}, vol.~9, no.~3,
  pp. 227--238, 2016.

\bibitem{pereira2015framework}
R.~d.~P. Pereira and P.~M. Engel, ``{A Framework for Constrained and Adaptive
  Behavior-based Agents},'' \emph{arXiv Preprint arXiv:1506.02312}, 2015.

\bibitem{dey2013ql}
R.~Dey and C.~Child, ``Ql-bt: Enhancing behaviour tree design and
  implementation with q-learning,'' in \emph{2013 IEEE Conference on
  Computational Intelligence in Games (CIG)}.\hskip 1em plus 0.5em minus
  0.4em\relax IEEE, 2013, pp. 1--8.

\bibitem{fu2016reinforcement}
Y.~Fu, L.~Qin, and Q.~Yin, ``A reinforcement learning behavior tree framework
  for game ai,'' in \emph{2016 International Conference on Economics, Social
  Science, Arts, Education and Management Engineering}.\hskip 1em plus 0.5em
  minus 0.4em\relax Atlantis Press, 2016.

\bibitem{french2019demonstration}
French, Wu, Pan, Zhou, and Jenkins, ``{Learning Behavior Trees From
  Demonstration},'' in \emph{Robotics and Automation (ICRA), 2019 IEEE
  International Conference on}.\hskip 1em plus 0.5em minus 0.4em\relax IEEE,
  2019.

\bibitem{sagredo2017trained}
I.~Sagredo-Olivenza, P.~P. G{\'o}mez-Mart{\'\i}n, M.~A. G{\'o}mez-Mart{\'\i}n,
  and P.~A. Gonz{\'a}lez-Calero, ``Trained behavior trees: Programming by
  demonstration to support ai game designers,'' \emph{IEEE Transactions on
  Games}, 2017.

\bibitem{paxton2016costar}
C.~Paxton, A.~Hundt, F.~Jonathan, K.~Guerin, and G.~D. Hager, ``{CoSTAR:
  Instructing Collaborative Robots with Behavior Trees and Vision},'' in
  \emph{Robotics and Automation (ICRA), 2017 IEEE International Conference
  on}.\hskip 1em plus 0.5em minus 0.4em\relax IEEE, 2017, pp. 564--571.

\bibitem{burridge1999sequential}
R.~R. Burridge, A.~A. Rizzi, and D.~E. Koditschek, ``Sequential {C}omposition
  of {D}ynamically {D}exterous {R}obot {B}ehaviors,'' \emph{The International
  Journal of Robotics Research}, vol.~18, no.~6, pp. 534--555, 1999.

\bibitem{conner2003composition}
D.~Conner, A.~Rizzi, and H.~Choset, ``Composition of local potential functions
  for global robot control and navigation,'' in \emph{IEEE/RSJ International
  Conference on Intelligent Robots and Systems, (IROS)}, vol.~4.\hskip 1em plus
  0.5em minus 0.4em\relax IEEE, pp. 3546--3551.

\bibitem{SRINIVASAN2002133}
\BIBentryALTinterwordspacing
B.~Srinivasan, P.~Huguenin, K.~Guemghar, and D.~Bonvin, ``A global
  stabilization strategy for an inverted pendulum,'' \emph{IFAC Proceedings
  Volumes}, vol.~35, no.~1, pp. 133 -- 138, 2002, 15th IFAC World Congress.
  [Online]. Available:
  \url{http://www.sciencedirect.com/science/article/pii/S1474667015386936}
\BIBentrySTDinterwordspacing

\bibitem{sprague2019learning}
C.~I. Sprague, D.~Izzo, and P.~{\"O}gren, ``Learning dynamic-objective policies
  from a class of optimal trajectories,'' in \emph{2020 59th IEEE Conference on
  Decision and Control (CDC)}.\hskip 1em plus 0.5em minus 0.4em\relax IEEE,
  2020, pp. 597--602.

\end{thebibliography}

\end{document}